\documentclass[journal]{IEEEtran}
\ifCLASSINFOpdf
\else
\fi

\usepackage{amsmath,amssymb,amsfonts,amsthm,amsbsy}
\usepackage{cite}
\usepackage{graphicx}
\usepackage{subfigure}
\usepackage{booktabs}
\usepackage{multirow}
\usepackage{multicol}
\usepackage{psfrag}
\usepackage{algorithm}
\usepackage{algorithmic}
\usepackage{enumitem}
\usepackage{hyperref}

\theoremstyle{remark}
\newtheorem{remark}{Remark}

\newcommand{\round}{\operatorname{round}}
\newcommand{\clip}{\operatorname{clip}}
\newcommand{\sign}{\operatorname{sign}}
\newcommand{\citet}{\cite}
\newcommand{\citep}{\cite}

\begin{document}
%
\title{Learning Sparse Low-Precision Neural Networks With Learnable Regularization}
%
%
%

\author{Yoojin Choi,
        Mostafa El-Khamy,~\IEEEmembership{Senior Member, IEEE,}
        Jungwon Lee,~\IEEEmembership{Fellow, IEEE}%
\thanks{Y. Choi, M. El-Khamy, and J. Lee are with the SoC R\&D, Samsung Semiconductor Inc., San Diego, CA 92121 USA (e-mail: yoojin.c@samsung.com; mostafa.e@samsung.com; jungwon2.lee@samsung.com).}}

\maketitle

\begin{abstract}
We consider learning deep neural networks (DNNs) that consist of low-precision weights and activations for efficient inference of fixed-point operations. In training low-precision networks, gradient descent in the backward pass is performed with high-precision weights while quantized low-precision weights and activations are used in the forward pass to calculate the loss function for training. Thus, the gradient descent becomes suboptimal, and accuracy loss follows. In order to reduce the mismatch in the forward and backward passes, we utilize mean squared quantization error (MSQE) regularization. In particular, we propose using a learnable regularization coefficient with the MSQE regularizer to reinforce the convergence of high-precision weights to their quantized values. We also investigate how partial L2 regularization can be employed for weight pruning in a similar manner. Finally, combining weight pruning, quantization, and entropy coding, we establish a low-precision DNN compression pipeline. In our experiments, the proposed method yields low-precision MobileNet and ShuffleNet models on ImageNet classification with the state-of-the-art compression ratios of 7.13 and 6.79, respectively. Moreover, we examine our method for image super resolution networks to produce 8-bit low-precision models at negligible performance loss.
\end{abstract}


%
\IEEEpeerreviewmaketitle

\section{Introduction} \label{sec:intro}

Deep neural networks (DNNs) have achieved performance breakthroughs in many of computer vision tasks~\citep{lecun2015deep}. The revolutionary progress of deep learning comes with over-parametrized multi-layer network architectures, and nowadays millions or tens of millions parameters in more than one hundred layers are not exceptional anymore. Network compression for efficient inference is of great interest for deployment of large-size DNNs on resource-limited platforms such as battery-powered mobile devices~\citep{sze2017efficient,cheng2018model}. In such resource-constrained hardware, not only memory and power are limited but also basic floating-point arithmetic operations are in some cases not supported. Hence, it is preferred and sometimes necessary to deliver compressed DNNs of low-precision fixed-point weights and activations (feature maps).

\begin{figure}[t!]
\centering
\includegraphics[width=\columnwidth]{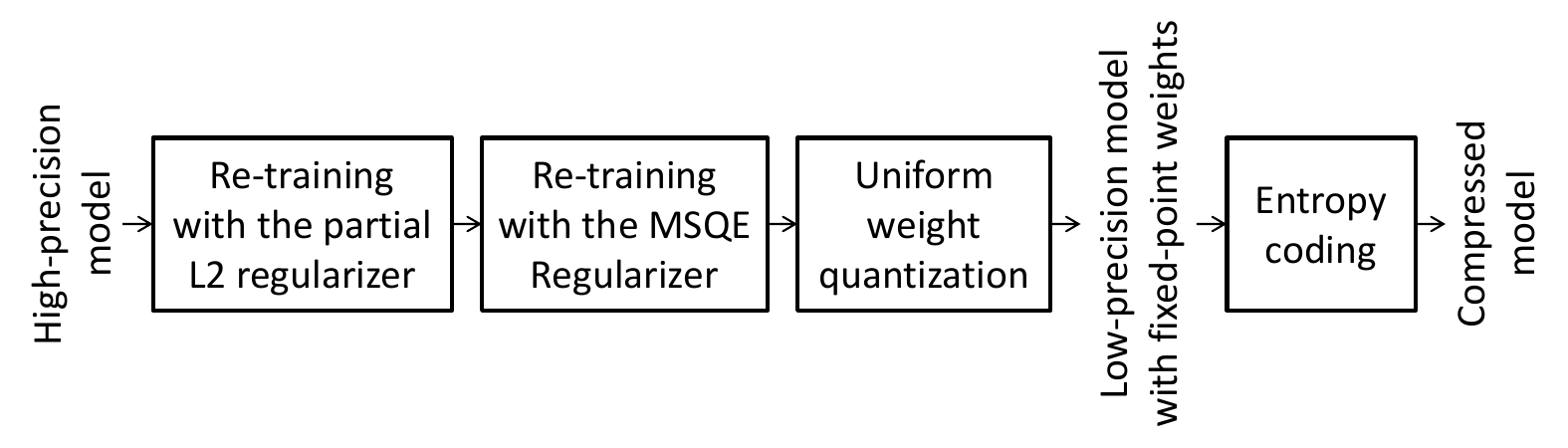}
\caption{Our low-precision DNN compression pipeline. We utilize partial L2 regularization and MSQE regularization to transform a pre-trained high-precision model into a sparse low-precision model with fixed-point weights and activations. The low-precision weights are further compressed in size with lossless entropy source coding.\label{sec:intro:fig:01}}
\end{figure}

In this paper, we propose a network compression scheme that produces sparse low-precision DNNs through learning with regularization. In particular, we let the regularization coefficient be learnable, instead of treating it as a fixed hyper-parameter, to make a smooth and efficient transition of a high-precision model into a sparse quantized model. The proposed compression pipeline is summarized in Figure~\ref{sec:intro:fig:01}.
\begin{itemize}
\item For weight pruning, we utilize partial L2 regularization to make a portion of small-value weights tend to zero so we can safely prune them at negligible accuracy loss.
\item For weight quantization, we regularize (unpruned) weights with another regularization term of the mean squared quantization error (MSQE). In this stage, we also quantize the activations (feature maps) of each layer to mimic low-precision operations at inference time. The quantization bin sizes for weights and activations are optimized to minimize their MSQEs in each layer.
\item The pruned and quantized model is converted into a low-precision model and its low-precision weights are further compressed in size with lossless entropy coding such as Huffman coding and universal source coding algorithms (e.g., see \citep[Section~11.3]{cover2012elements}) for memory-efficient deployment.
\end{itemize}

It is difficult to train low-precision DNNs with standard gradient descent since the learning rate is typically set to be a small floating-point value but low-precision weights cannot be adjusted in fine resolution. To enable training low-precision DNNs, a series of papers on binary neural networks suggests utilizing high-precision shadow weights to accumulate the negatives of the gradients in fine resolution, while the gradients are obtained from the network loss function calculated with binarized (or quantized) weights~\citep{courbariaux2015binaryconnect,lin2016neural,hubara2016binarized}. That is, high-precision weights are quantized in the forward pass, but the quantization function is replaced with the identity function in the backward pass for gradient descent. This approximate gradient descent algorithm is further refined in the subsequent works~\citep{rastegari2016xnor,zhou2016dorefa,zhu2017trained,cai2017deep,hou2017loss,hou2018loss,gysel2018ristretto,zhou2018explicit}.

BinaryRelax~\cite{yin2018binaryrelax} proposed relaxation of the quantization problem via Moreau envelope (also known as Moreau-Yosida regularization)~\cite{moreau1965proximite,yosida1965functional} and used pseudo quantized weights in the forward pass to solve the relaxed quantization problem. In particular, the pseudo quantized weights are obtained by weighted average of high-precision weights and their quantized values. By manually adjusting the weighting factor in the weighted average, the pseudo quantized weights are pushed towards their quantized values gradually in training. In \cite{yin2019blended}, the blended coarse gradient descent (BCGD) algorithm was proposed, where the BinaryConnect scheme~\cite{courbariaux2015binaryconnect} and the standard projected gradient descent algorithm (PGD)~\cite{combettes2016stochastic} are combined with some blending parameter. For quantization of activations, parameterized clipping activation (PACT)~\cite{choi2018pact} proposed using an activation clipping parameter that is optimized during training to find the right quantization scale. The two-valued proxy derivative of the parametric activation function in \cite{choi2018pact} was further enhanced by three-valued proxy partial derivative in \cite{yin2019blended}. LQ-Nets~\cite{zhang2018lq} proposed finding optimal quantization levels in a subspace compatible with bit-wise operations. In \cite{faraone2018syq}, it was proposed to learn separate scaling factors for fine-grained weight subgroups (e.g., pixel-wise or row-wise scaling factors).

The mismatch in the forward and backward passes results in sub-optimal gradient descent that causes accuracy loss. The mismatch is more problematic for the models using lower-precision weights and activations, since the quantization error is more significant. There have been some attempts to reduce this mismatch by introducing better backward pass approximation, e.g., using clipped ReLU and log-tailed ReLU instead of the linear function (e.g., see \citep{cai2017deep}). Recently, it was proposed to use smooth differentiable approximation of the staircase quantization function. In \cite{liu2019learning}, affine combination of high-precision weights and their quantized values, called alpha blending, was used to replace the quantization function. In \cite{yang2019quantization}, the quantization function was approximated as a linear combination of several sigmoid functions with learnable biases and scales. Similarly, differentiable soft quantization (DSQ)~\cite{gong2019differentiable} exploited a series of hyperbolic tangent functions to approximate the staircase quantization function. The proposed approximation gradually approaches to the quantization function in training by adjusting the blending factor or the temperature parameter in the sigmoid function. Our approach is different from these efforts. We use regularization to steer high-precision weights to converge to their quantized values so that the mismatch between high-precision weights and quantized weights becomes smaller instead of enhancing the backward pass approximation.

We reduce the mismatch between high-precision weights and quantized weights with MSQE regularization. In particular, we propose making the regularization coefficient learnable. Using learnable regularization, high-precision weights are reinforced to converge to their quantized values gradually in training. We empirically show that our learnable regularization yields more accurate low-precision models than the conventional regularization with a fixed regularization coefficient. MSQE is a well-known distortion metric in data quantization, and it has been used in network quantization as well to reduce the performance loss from quantization (e.g., see \cite{anwar2015fixed,rastegari2016xnor}). Our contribution is to use MSQE as a regularizer with a learnable coefficient, which is new to the best of our knowledge. The loss-aware weight quantization in \citet{hou2017loss,hou2018loss} proposed the proximal Newton algorithm to minimize the loss function under the constraints of low-precision weights, which is however impractical for large-size networks due to the prohibitive computational cost to estimate the Hessian matrix of the loss function. Our method simply uses the stochastic gradient descent, while the mismatch between high-precision weights and quantized weights is minimized with the MSQE regularization. No regularization is considered in \citet{hou2017loss,hou2018loss}. Relaxed quantization~\citet{louizos2018relaxed} introduced a differentiable quantization procedure by transforming continuous distributions of weights and activations to differentiable soft categorical distributions. Our method is much simpler than the relaxation procedure in \citet{louizos2018relaxed}, since it only requires MSQE regularization. Furthermore, it shows better performance than \citet{louizos2018relaxed} empirically in MobileNet quantization.

Weight pruning curtails redundant weights completely from DNNs so one can skip the computations for pruned weights. Some successful pruning algorithms can be found in \citet{han2015learning,lebedev2016fast,wen2016learning,guo2016dynamic,lin2017runtime}. In this paper, we discuss how partial L2 regularization can be used for weight pruning. Finally, combining weight pruning, quantization, and entropy coding, as shown in Figure~\ref{sec:intro:fig:01}, we achieve the state-of-the-art compression results for low-precision MobileNet~\citep{howard2017mobilenets} and ShuffleNet~\citep{zhang2018shufflenet} on ImageNet classification.

Weight sharing is another network compression scheme studied in \citet{han2015deep,choi2017towards,ullrich2017soft,molchanov2017variational,agustsson2017soft,louizos2017bayesian,tung2018deep,choi2020universal}. It reduces the number of distinct weight values in DNNs by quantization. In contrast to low-precision weights from uniform quantization, weight sharing allows non-uniform quantization. For non-uniform quantization (e.g., $k$-means clustering), quantization output levels (e.g., cluster centers) do not have to be evenly spaced, and they are usually high-precision floating-point values. The quantization output levels are the shared weight values used in inference. Thus, floating-point arithmetic operations are still needed in inference, although the quantized weights can be compressed in size by lossless source coding (e.g., Huffman coding).

We finally note that reinforcement learning has been proposed as a promising methodology to search for quantized and/or compressed models that satisfy certain latency, energy, and/or model size requirements, given hardware specifications to deploy the models~\citep{he2018amc,wang2019haq}.

\section{Low-precision DNN model} \label{sec:model}

We consider low-precision DNNs that are capable of efficient processing in the inference stage by using fixed-point arithmetic operations. In particular, we focus on the fixed-point implementation of convolutional and fully-connected layers, since they are the dominant parts of computational costs and memory requirements in DNNs (see \citep[Table~II]{sze2017efficient}).

The major bottleneck of efficient DNN processing is known to be in memory accesses~\citep[Section~V-B]{sze2017efficient}. Horowitz provides rough energy costs of various arithmetic and memory access operations for 45 nm technology~\citep[Figure~1.1.9]{horowitz20141}, where we can find that memory accesses typically consume more energy than arithmetic operations, and the memory access cost increases with the read size. Hence, for example, deploying binary models, instead of 32-bit models, it is expected to reduce energy consumption by $32\times$ at least, due to $32$ times fewer memory accesses.

Low-precision weights and activations basically stem from uniform quantization (e.g., see \citep[Section~5.4]{gersho2012vector}), where quantization bin boundaries are uniformly spaced and quantization output levels are the midpoints of bin intervals. Quantized weights and activations are represented by fixed-point numbers of small bit-width. Scaling factors (i.e., quantization bin sizes) are defined in each layer for fixed-point weights and activations, respectively, to alter their dynamic ranges.

\begin{figure}[!t]
\centering
\includegraphics[width=\columnwidth]{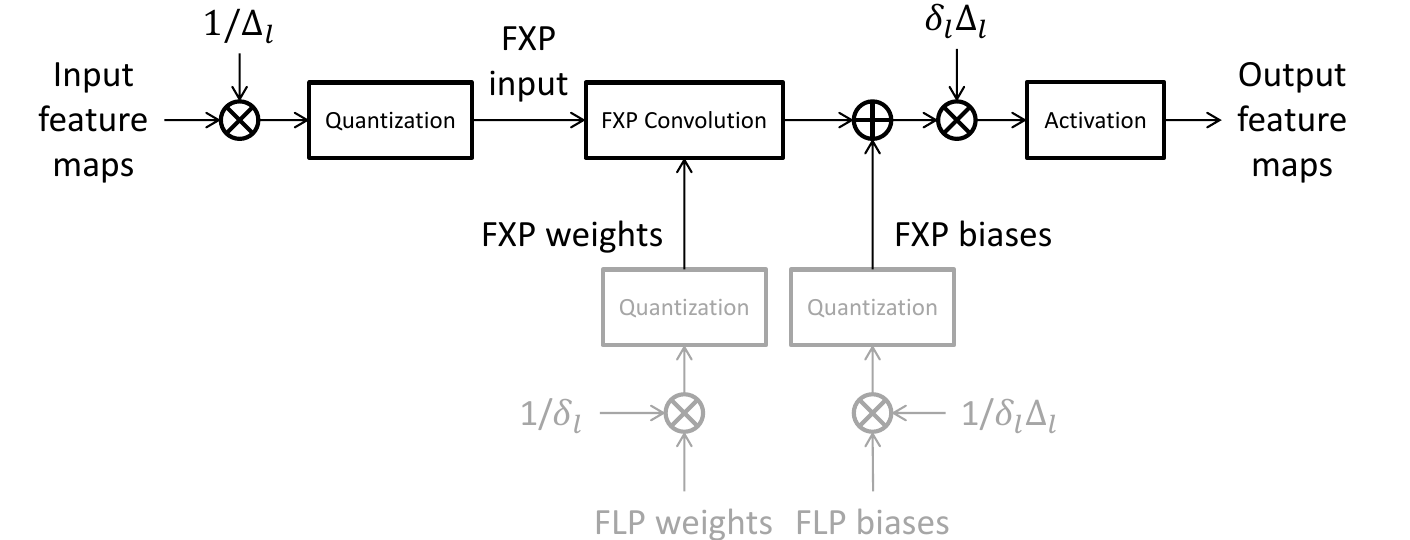}
\caption{Low-precision convolutional layer using fixed-point (FXP) convolution and bias addition.\label{sec:model:fig:01}}
\end{figure}

Figure~\ref{sec:model:fig:01} shows the fixed-point design of a general convolutional layer consisting of convolution, bias addition and non-uniform activation. Fixed-point weights and input feature maps are given with common scaling factors~$\delta_l$ and $\Delta_l$, respectively, where $l$ is the layer index. Then, the convolution operation can be implemented by fixed-point multipliers and accumulators. Biases are added, if present, after the convolution, and then the output is scaled properly by the product of the scaling factors for weights and input feature maps, i.e., $\delta_l\Delta_l$, as shown in the figure. Here, the scaling factor for the biases is specially set to be $\delta_l\Delta_l$ so that fixed-point bias addition can be done easily without another scaling. Then, a non-linear activation function follows. Finally, the output activations are fed into the next layer as the input. 

Using rectified linear unit (ReLU) activation, two scaling operations across two layers, i.e., scaling operations by $\delta_l\Delta_l$ and $1/\Delta_{l+1}$, can be combined into one scaling operation by $\delta_l\Delta_l/\Delta_{l+1}$ before (or after) ReLU activation. Furthermore, if the scaling factors are power-of-two numbers, then one can even implement scaling by bit-shift. Similarly, low-precision fully-connected layers can be implemented by replacing convolution with matrix multiplication in the figure.

\section{Regularization for Low-precision DNNs} \label{sec:reg}

In this section, we present the regularizers that are utilized to learn pruned and quantized DNNs of low-precision weights and activations. We first define the quantization function. Given the number of bits, i.e., bit-width~$n$, the quantization function yields
\begin{equation} \label{sec:reg:qfunc:eq:01}
Q_n(x;\delta)
=\begin{cases}
\delta\clip_n(\round(x/\delta)), & n\geq2, \\
\delta\sign(x), & n=1,
\end{cases}
\end{equation}
where $x$ is the input and $\delta$ is the scaling factor; we let
\[\begin{split}
\round(x)&=\sign(x)\lfloor|x|+0.5\rfloor, \\
\clip_n(x)&=\min(\max(x,-2^{n-1}),2^{n-1}-1),
\end{split}\]
where $\lfloor x \rfloor$ is the largest integer smaller than or equal to $x$. For ReLU activation, the ReLU output is always non-negative, and thus we use the unsigned quantization function given by
\begin{equation} \label{sec:reg:qfunc:eq:02}
Q_n^+(x;\delta)
=\delta\clip_n^+(\round(x/\delta)),
\end{equation}
for $n\geq1$, where $\clip_n^+(x)=\min(\max(x,0),2^n-1)$.

\subsection{Regularization for weight quantization} \label{sec:reg:weightquant}

\begin{figure*}[t!]
\centering
\subfigure[Iterations=$10$k]{\includegraphics[width=0.22\textwidth]{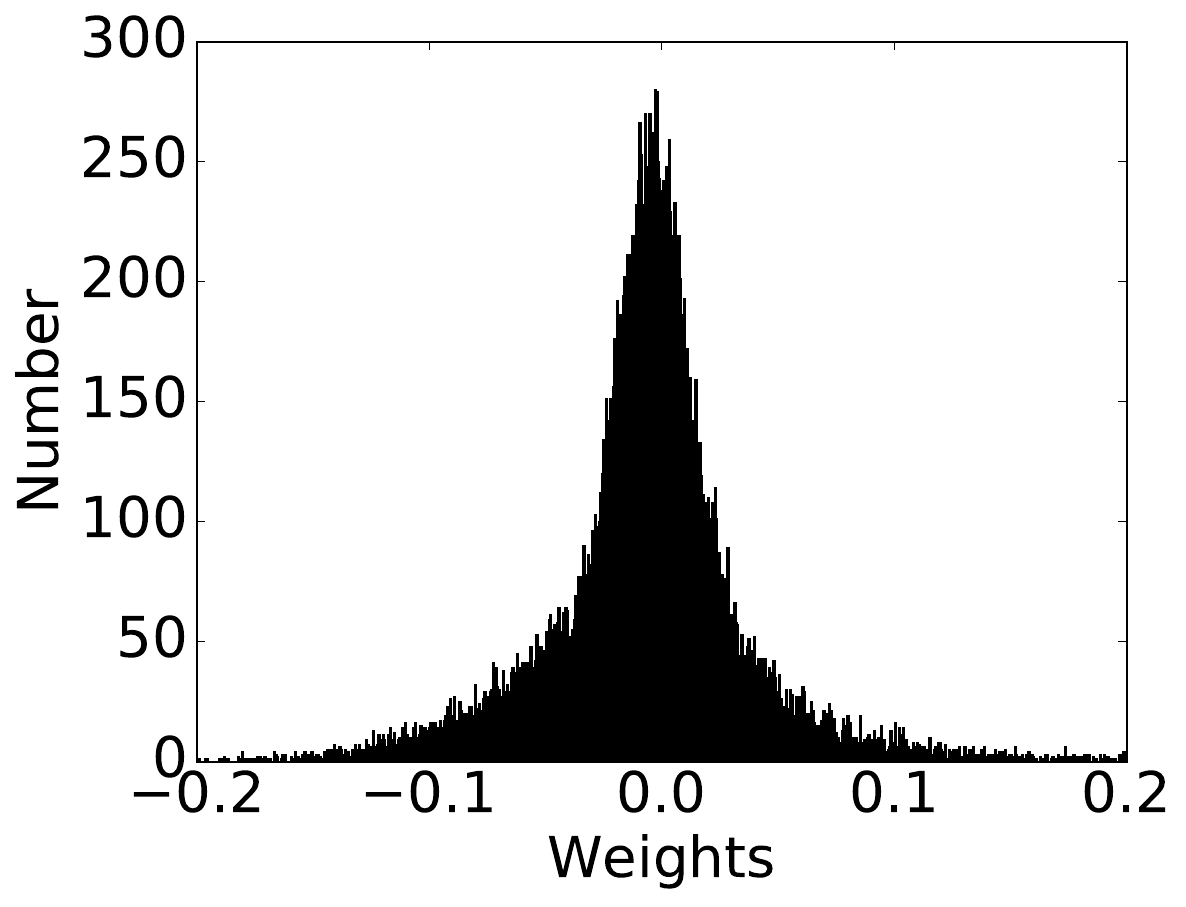}}
\subfigure[Iterations=$21$k]{\includegraphics[width=0.22\textwidth]{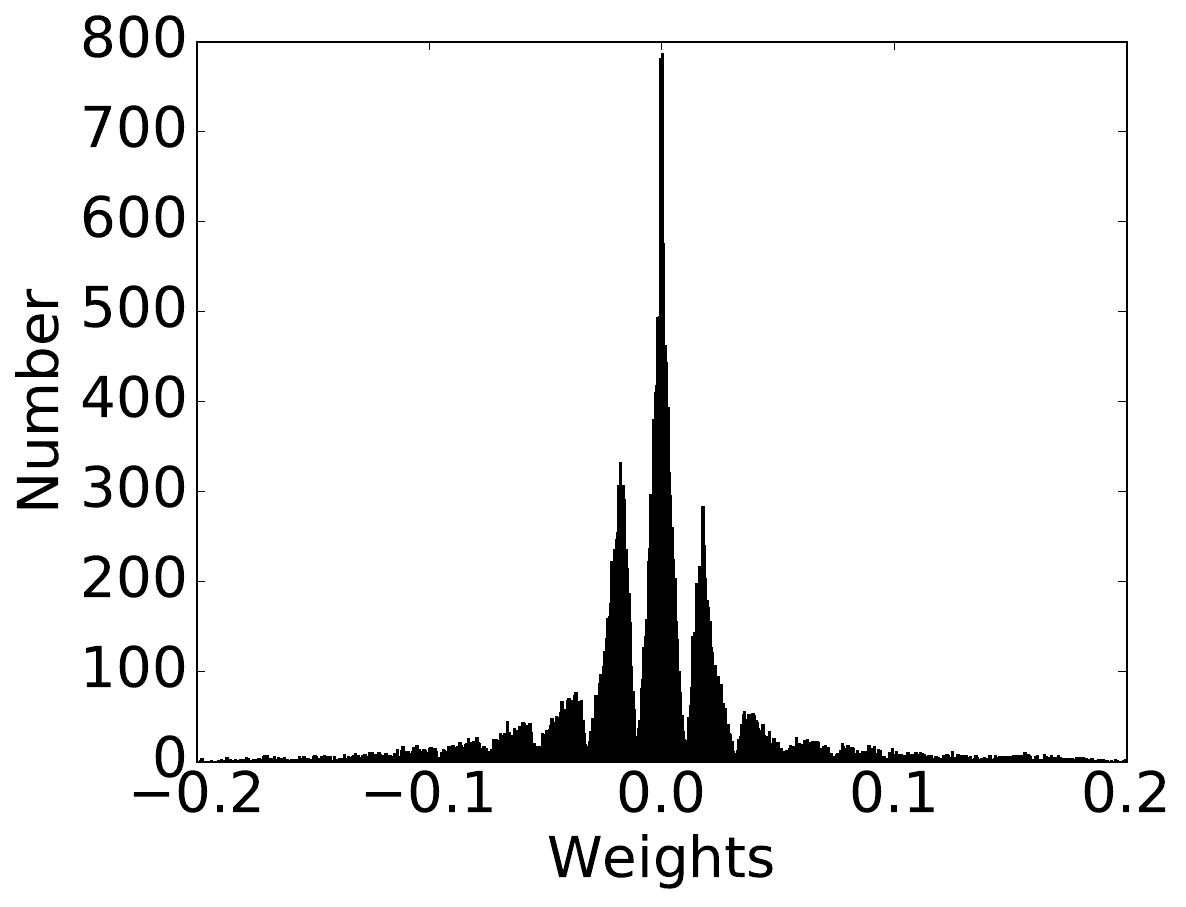}}
\subfigure[Iterations=$23$k]{\includegraphics[width=0.22\textwidth]{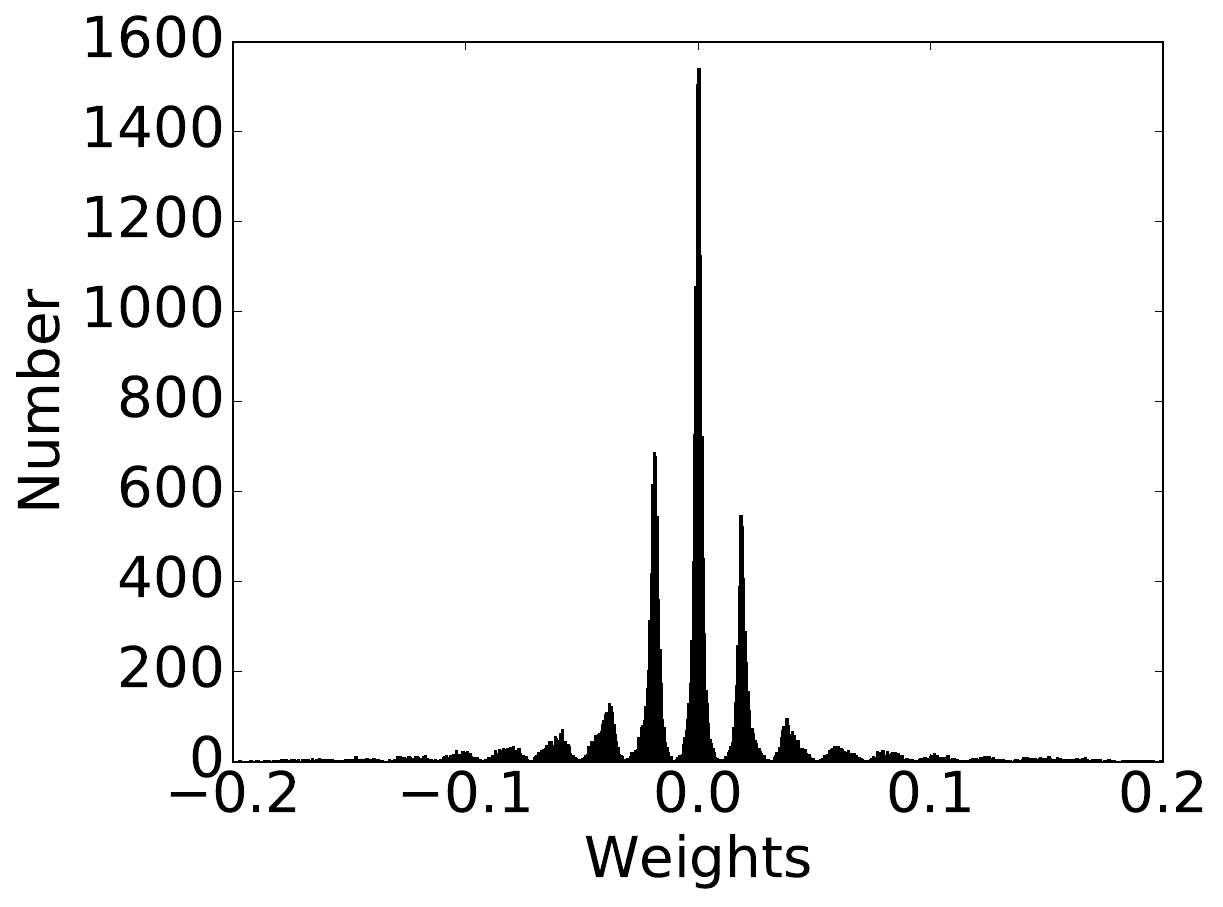}}
\subfigure[Iterations=$30$k]{\includegraphics[width=0.22\textwidth]{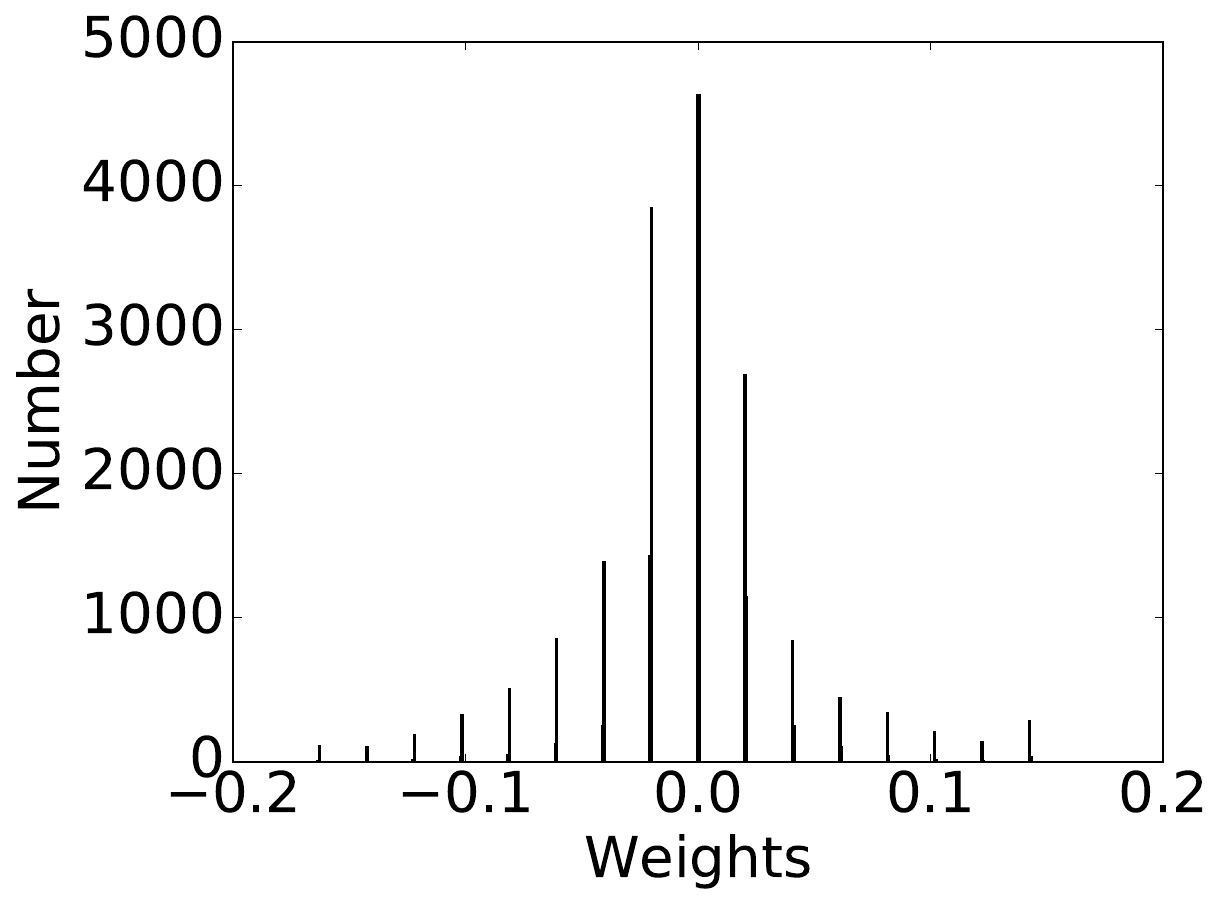}}
\caption{Weight histogram snapshots of the MNIST LeNet-5 second convolutional layer captured at different training batch iteration numbers while a pre-trained model is quantized to have 4-bit weights and activations with the proposed regularization method.\label{sec:reg:weightquant:fig:01}}
\end{figure*}

Consider a general non-linear neural network consisting of $L$ layers. Let $\mathcal{W}_1,\mathcal{W}_2,\dots,\mathcal{W}_L$ be the sets of high-precision weights in layers $1$ to $L$, respectively. For notational simplicity, we let $A_1^L=A_1,A_2,\dots,A_L$ for any symbol~$A$. We define the MSQE regularizer for weights of all $L$ layers as
\begin{equation} \label{sec:reg:weightquant:eq:01}
R_n(\mathcal{W}_1^L;\delta_1^L)
=\frac{1}{N}\sum_{l=1}^L\sum_{w\in\mathcal{W}_l}|w-Q_n(w;\delta_l)|^2,
\end{equation}
where $n$ is the bit-width for quantized weights, $\delta_l$ is the scaling factor (i.e., quantization bin size) for quantized weights, and $N$ is the total number of weights from all layers, i.e.,
\[
N=\sum_{l=1}^L|\mathcal{W}_l|,
\]
where $|\mathcal{W}_l|$ is the number of weights in layer~$l$. We assumed that bit-width~$n$ is the same for all layers, just for notational simplicity, but it can be easily extended to more general cases such that each layer has a different bit-width.

Including the MSQE regularizer in \eqref{sec:reg:weightquant:eq:01}, the cost function to optimize in training is given by
\begin{multline} \label{sec:reg:weightquant:eq:02}
C_n(\mathcal{X};\mathcal{W}_1^L,\delta_1^L) \\
=E(\mathcal{X};Q_n(\mathcal{W}_1^L;\delta_1^L))+\lambda R_n(\mathcal{W}_1^L;\delta_1^L),
\ \ \
\lambda>0,
\end{multline}
where, with a slight abuse of notation, $Q_n(\mathcal{W}_1^L;\delta_1^L)$ denotes the set of quantized weights of all $L$ layers, $E(\mathcal{X};Q_n(\mathcal{W}_1^L))$ is the target loss function evaluated on the training dataset~$\mathcal{X}$ using the quantized weights, and $\lambda$ is the regularization coefficient. We set the scaling factors~$\delta_1^L$ to be learnable parameters and optimize them along with weights~$\mathcal{W}_1^L$.

\begin{remark} \label{sec:reg:weightquant:remark:01}
We clarify that we use high-precision weights in the backward pass for gradient descent by replacing approximately the quantization function $Q_n$ with the identity function. In the forward pass, we use quantized weights and activations, and the target objective function~$E$ is also calculated with the quantized weights and activations to mimic the low-precision inference-stage loss. Hence, the final trained models are low-precision models, which can be operated on low-precision fixed-point hardware in inference with no accuracy loss. Note that our method still has the gradient mismatch problem, similar to the existing approaches (see Section~\ref{sec:intro}). However, by adding the MSQE regularizer, we encourage high-precision weights to converge to their quantized values so that we reduce the mismatch.
\end{remark}

\textbf{Learnable regularization coefficient}. The regularization coefficient~$\lambda$ in \eqref{sec:reg:weightquant:eq:02} is a hyper-parameter that controls the trade-off between the loss and the regularization. It is conventionally fixed ahead of training. However, searching for a good hyper-parameter value is usually time-consuming. Hence, we propose the learnable regularization coefficient, i.e., we let the regularization coefficient be another learnable parameter.

We start training with a small initial value for $\lambda$, i.e., with little regularization. However, we promote the increase of $\lambda$ in training by adding a penalty term for a small regularization coefficient, which is $-\log{\lambda}$ for $\lambda>0$, in the cost function (see \eqref{sec:reg:weightquant:eq:03}). The increasing coefficient~$\lambda$ reinforces the convergence of high-precision weights to their quantized values for reducing the MSQE. It consequently alleviates the gradient mismatch problem (see Remark~\ref{sec:reg:weightquant:remark:01}). The cost function in \eqref{sec:reg:weightquant:eq:02} is altered into
\begin{multline} \label{sec:reg:weightquant:eq:03}
C_n(\mathcal{X};\mathcal{W}_1^L,\delta_1^L,\lambda) \\
=E(\mathcal{X};Q_n(\mathcal{W}_1^L;\delta_1^L))
+\lambda R_n(\mathcal{W}_1^L;\delta_1^L)-\log{\lambda}.
\end{multline}
For gradient descent, we need the gradients of \eqref{sec:reg:weightquant:eq:03} with respect to weights, scaling factors and the regularization coefficient, respectively, which are provided in Appendix.

\begin{remark}
In \eqref{sec:reg:weightquant:eq:03}, observe that we use quantized weights in the forward pass to compute the loss while we update high-precision weights with gradient descent in the backward pass, as in BinaryConnect~\cite{courbariaux2015binaryconnect}. Thus, our method is different from BinaryRelax~\cite{yin2018binaryrelax} that uses pseudo quantized weights in the forward pass. The pseudo quantized weights are computed by weighted average of high-precision weights and their quantized values. Our MSQE regularization resembles Moreau-Yosida regularization in BinaryRelax. However, the Moreau-Yosida regularization factor in BinaryRelax is manually increased with a fixed rate at every iteration in training so the pseudo quantized weights are pushed towards quantized values as training goes on. In our scheme, the difference between high-precision weights and quantized weights is reduced by the MSQE regularization. Moreover, we propose letting the regularization coefficient~$\lambda$ be learnable and adding another penalty term $-\log\lambda$ to promote increasing $\lambda$; hence, $\lambda$ does not necessarily increase with a fixed rate and can saturate after some point of training to find a better local optimum, as shown in Figure~\ref{sec:exp:res:resnet18:fig:02}(a). We do not constrain the range of $\lambda$ in \eqref{sec:reg:weightquant:eq:03} so it is possible that $\lambda$ diverges in optimization. However, we empirically found that $\lambda$ saturates after some point of training in practice as the loss saturates (e.g., see Figure~\ref{sec:exp:res:resnet18:fig:02}(a)).
\end{remark}

\textbf{Evolution of weight histogram}. Figure~\ref{sec:reg:weightquant:fig:01} presents an example of how high-precision weights are gradually quantized by our regularization scheme. We plotted weight histogram snapshots captured at the second convolutional layer of the MNIST LeNet-5 model\footnote{\url{https://github.com/BVLC/caffe/tree/master/examples/mnist}} 
while a pre-trained model is quantized to a 4-bit fixed-point model. The histograms in the figure from the left to the right correspond to $10$k, $21$k, $23$k, and $30$k batch iterations in training, respectively. Observe that the weight distribution gradually converges to the sum of uniformly spaced delta functions and all high-precision weights converge to quantized values completely in the end.

\textbf{Comparison to soft weight sharing}. In soft weight sharing~\citep{nowlan1992simplifying,ullrich2017soft}, a Gaussian mixture prior is assumed, and the model is regularized to form groups of weights that have similar values around the Gaussian component centers (e.g., see \citep[Section~5.5.7]{bishop2006pattern}). The learnable regularization coefficient can be related to the learnable variance in the Gaussian mixture prior. However, our weight regularization method is different from soft weight sharing since we consider uniform quantization and optimize quantization bin sizes, instead of optimizing individual Gaussian component centers for non-uniform quantization. We employ the simple MSQE regularization term for quantization, so that it is applicable to large-size DNNs. Note that soft weight sharing yields the regularization term of the logarithm of the summation of exponential functions, which is sometimes too complex to compute for large-size DNNs. In our method, the additional computational complexity for MSQE regularization is not expensive. It only scales in the order of $O(N)$, where $N$ is the number of weights. Hence, the proposed scheme is easily applicable to the state-of-the-art DNNs with millions or tens of millions weights.

We note that biases are treated similar to weights. However, for the fixed-point design presented in Section~\ref{sec:model}, we use $\delta_l\Delta_l$ instead of $\delta_l$ as the scaling factor in \eqref{sec:reg:weightquant:eq:01}, where $\Delta_l$ is the scaling factor for input feature maps (i.e., activations from the previous layer), which is determined by the following activation quantization procedure.

\begin{figure*}[t!]
\centering
\subfigure[Iterations=$4$k]{\includegraphics[width=0.22\textwidth]{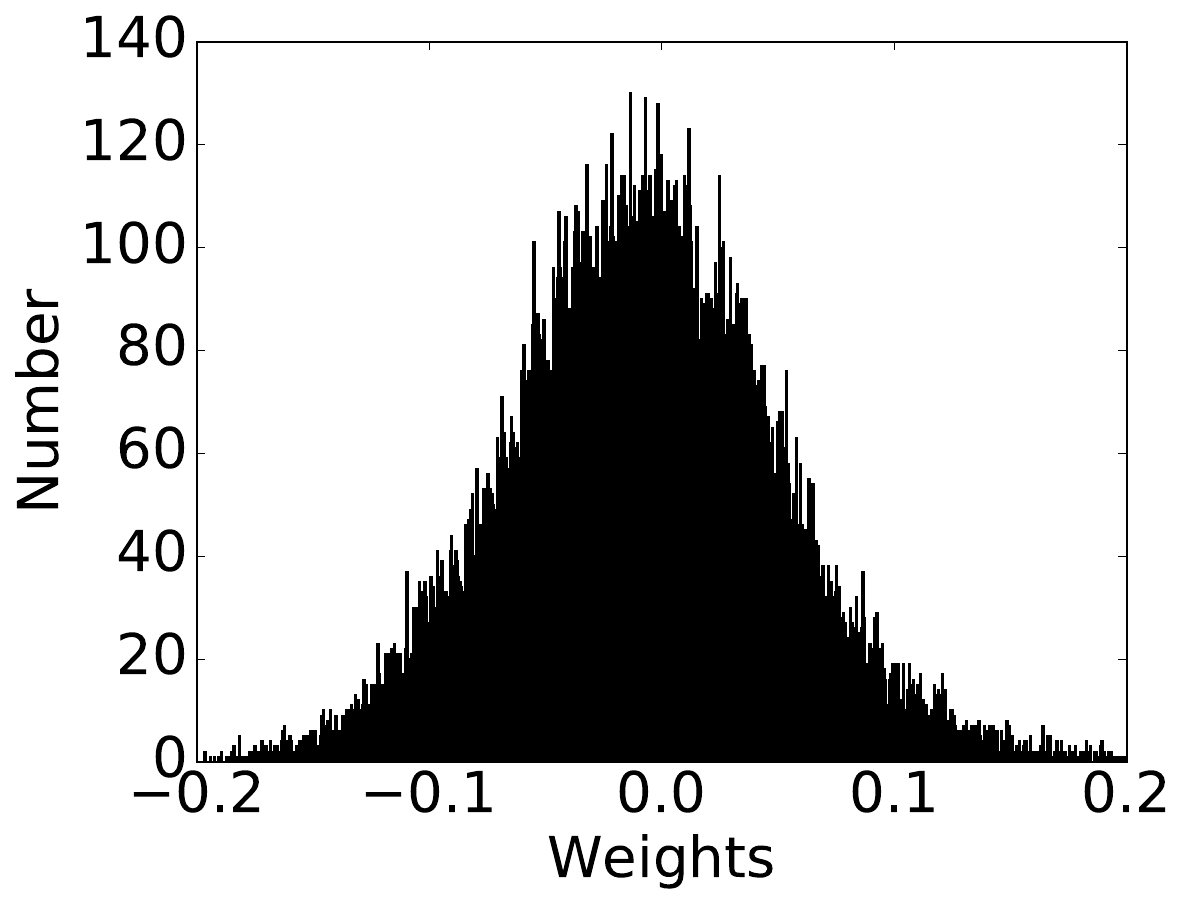}}
\subfigure[Iterations=$6$k]{\includegraphics[width=0.22\textwidth]{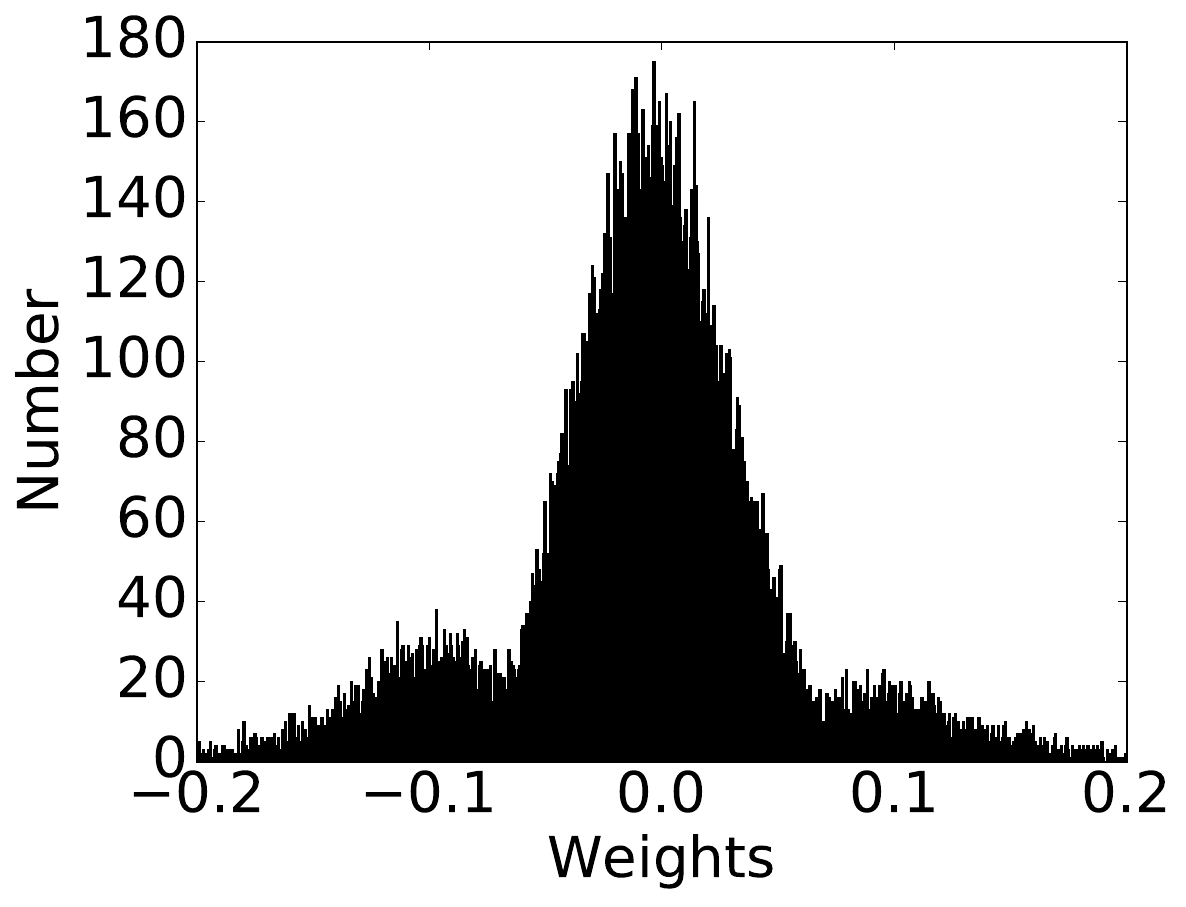}}
\subfigure[Iterations=$8$k]{\includegraphics[width=0.22\textwidth]{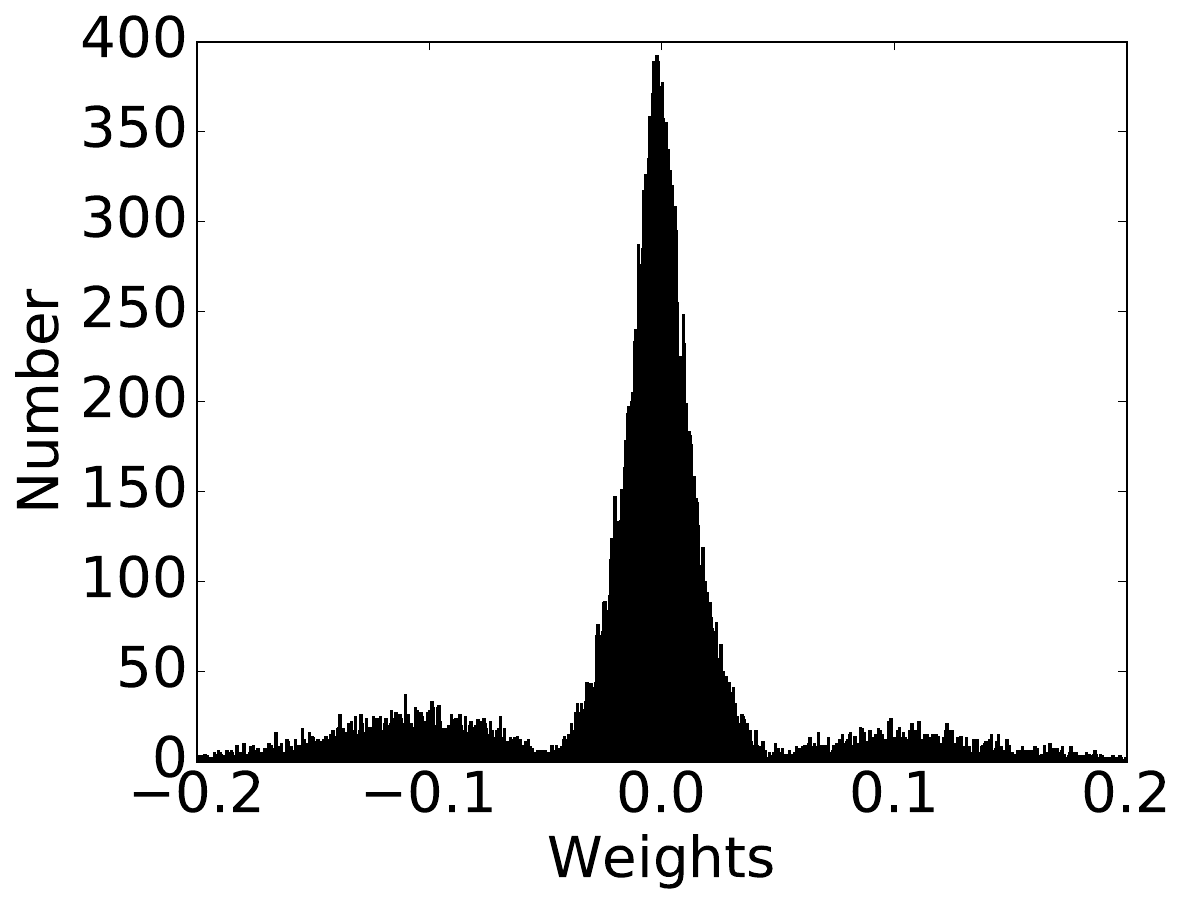}}
\subfigure[Iterations=$10$k]{\includegraphics[width=0.22\textwidth]{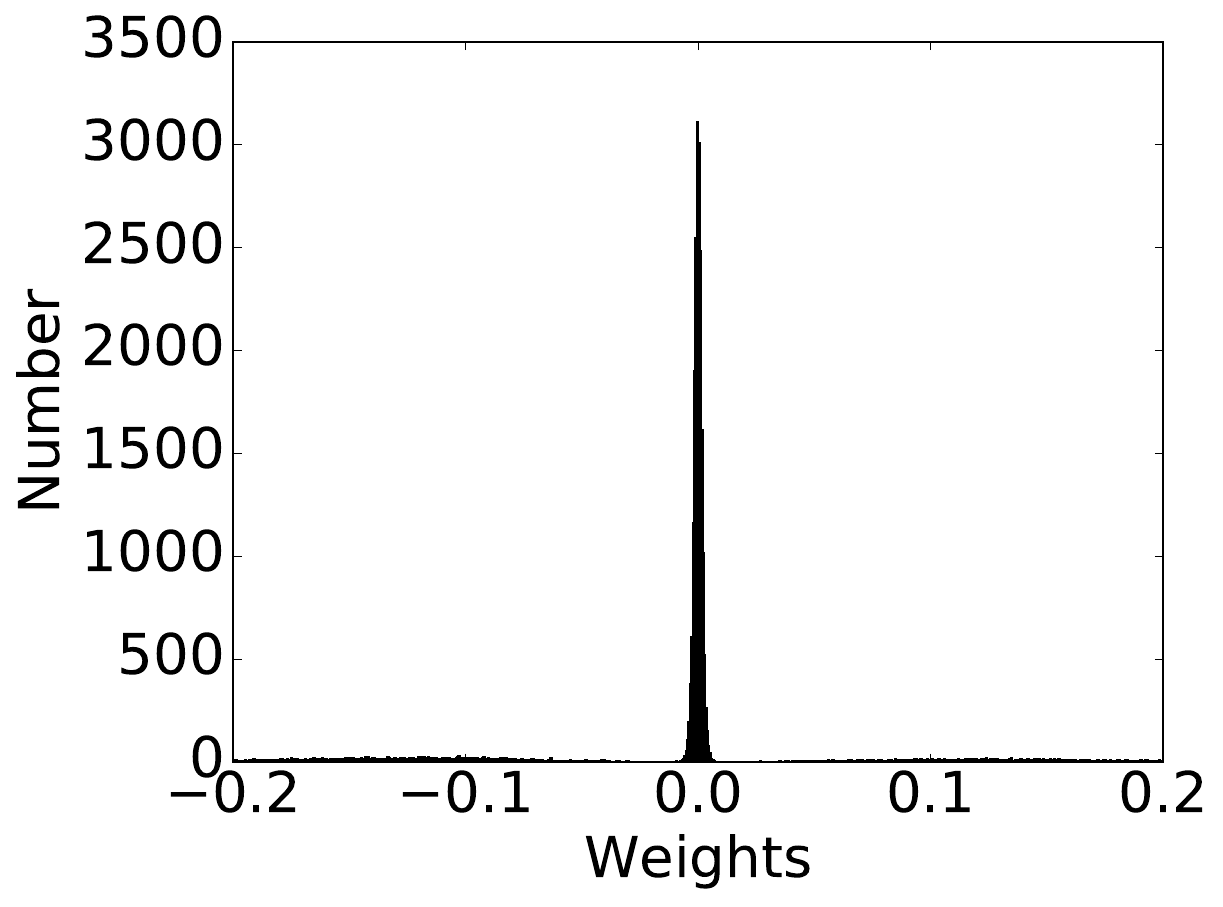}}
\caption{Weight histogram snapshots of the MNIST LeNet-5 at different training batch iteration numbers when trained from scratch with the partial L2 regularizer for 90\% sparsity ($r=90$).\label{sec:reg:pruning:fig:01}}
\end{figure*}

\subsection{Quantization of activations} \label{sec:reg:activation}

We quantize the output activation (feature map)~$x$ of layer~$l$ for $1\leq l\leq L$ and yield $Q_m^+(x;\Delta_l)$, where $Q_m^+$ is the quantization function in \eqref{sec:reg:qfunc:eq:02} for bit-width~$m$ and $\Delta_l$ is the learnable scaling factor for quantized activations of layer~$l$. We note that $\Delta_l$ is the scaling factor for activations of layer~$l$ whereas it denotes the scaling factor for input feature maps of layer~$l$ in Section~\ref{sec:model} (see Figure~\ref{sec:model:fig:01}). This is just one index shift in the notation, since the output of layer~$l$ is the input to layer~$l+1$. We adopt this change just for notational simplicity. Similar to \eqref{sec:reg:weightquant:eq:01}, we assumed that activation bit-width~$m$ is the same for all layers, but this constraint can be easily relaxed to cover the cases where each layer has a different bit-width. We assumed ReLU activation and used the unsigned quantization function~$Q_m^+$ while we can replace $Q_m^+$ with $Q_m$ in case of general non-linear activation (see \eqref{sec:reg:qfunc:eq:01} and \eqref{sec:reg:qfunc:eq:02}).

We optimize $\Delta_l$ by minimizing the MSQE for activations of layer~$l$, i.e., we minimize
\begin{equation} \label{sec:reg:activation:eq:01}
S_m(\mathcal{A}_l;\Delta_l)=\frac{1}{|\mathcal{A}_l|}\sum_{x\in\mathcal{A}_l}|x-Q_m^+(x;\Delta_l)|^2,
\end{equation}
where $\mathcal{A}_l$ is the set of activations of layer~$l$ for $1\leq l\leq L$. In the backward pass, we first perform gradient descent for weights and their scaling factors using the loss function in \eqref{sec:reg:weightquant:eq:03}, and then we update $\Delta_l$ with gradient descent using \eqref{sec:reg:activation:eq:01}. We do not utilize \eqref{sec:reg:activation:eq:01} in gradient descent for weights.

\textbf{Backpropagation through quantized activations}. Backpropagation is not feasible through quantized activations analytically since the gradient is zero almost everywhere. For backpropagation through the quantization function, we adopt the straight-through estimator~\citep{bengio2013estimating}. In particular, we pass the gradient through the quantization function when the input is within the clipping boundary. If the input is outside the clipping boundary, we pass zero.

\subsection{Regularization for weight pruning} \label{sec:reg:pruning}

For weight pruning, we propose using partial L2 regularization. In particular, given a target pruning ratio~$r$, we find the $r$-th percentile of weight magnitude values. Assuming that we prune the weights below this $r$-th percentile value in magnitude, we define a L2 regularizer only for them as follows:
\[
P_r(\mathcal{W}_1^L)
=\frac{1}{N}\sum_{l=1}^L\sum_{w\in\mathcal{W}_l}|w|^21_{|w|<\theta(r)},
\]
where $\theta(r)$ is the $r$-th percentile of weight magnitude values, which is the threshold for pruning. Adopting the learnable regularization coefficient as in \eqref{sec:reg:weightquant:eq:03}, we have
\[
C_r(\mathcal{X};\mathcal{W}_1^L,\lambda) \\
=E(\mathcal{X};\mathcal{W}_1^L)+\lambda P_r(\mathcal{W}_1^L)-\log{\lambda},
\]
for $\lambda>0$.

The partial L2 regularizer encourages the weights below the threshold to move towards zero, while the other unregularized weights are updated to minimize the loss due to pruning. The threshold~$\theta(r)$ is also updated at every iteration of training based on the instant weight distribution. We note that the threshold~$\theta(r)$ decreases as training goes on since the regularized weights gradually converge to zero (see Figure~\ref{sec:reg:pruning:fig:01}). After finishing the regularized training, we finally have a set of weights clustered very near zero. The loss due to pruning these small-value weights is negligible. 

After weight pruning, the pruned model is quantized by following the quantization procedure in Section~\ref{sec:reg:weightquant} and Section~\ref{sec:reg:activation}. In this stage, pruned weights are fixed to be zero while only unpruned weights are updated and quantized. After pruning, we still use quantization bins around zero for the weights that are not pruned but have small magnitude, or for the weights that are made to be small after training the quantized network; unpruned weights between $-\Delta/2$ to $\Delta/2$ are still quantized to zero, where $\Delta$ is the quantization bin size. However, the number of (unpruned) weights that are quantized to zero becomes much smaller after pruning.

\section{Experiments} \label{sec:exp}

We evaluate the proposed low-precision DNN compression for ImageNet classification and image super resolution. Image super resolution is included in our experiments as a regression problem since its accuracy is more sensitive to quantization than classification. Note that Tensorflow Lite\footnote{\url{https://www.tensorflow.org/lite}} already supports a very efficient 8-bit weight and activation quantization tool for network development on mobile platforms. Thus, our experimental results focus on more extreme cases of quantization using less than 8 bits, where a more sophisticated algorithm is needed for smaller loss. We use FLP and FXP to denote the floating-point and fixed-point formats, respectively.

\subsection{Experimental settings} \label{sec:exp:setting}

For ImageNet classification, we use the ImageNet ILSVRC 2012 dataset~\citep{russakovsky2015imagenet}. For image super resolution, we use the Open Images dataset\footnote{\url{https://github.com/openimages/dataset}} as the training dataset, which is pre-processed as described in \citet{ren2018ctsrcnn}. The proposed network pruning, quantization, and compression pipeline is implemented with Caffe\footnote{\url{https://github.com/BVLC/caffe}}. The pre-trained models used in our ImageNet classification experiments are obtained from the links in Table~\ref{sec:exp:setting:tbl:01}. For image super resolution, we train (CT-)SRCNNs from scratch as described in \citet{ren2018ctsrcnn}.

\setlength{\tabcolsep}{0.2em}
\begin{table}[t!]
\centering
\caption{Pre-trained models used in ImageNet classification experiments.\label{sec:exp:setting:tbl:01}}
{\scriptsize
\begin{tabular}{rl}
\toprule
AlexNet    & \url{https://github.com/BVLC/caffe/tree/master/models/bvlc_alexnet} \\
ResNet-18  & \url{https://github.com/HolmesShuan/ResNet-18-Caffemodel-on-ImageNet} \\
MobileNet  & \url{https://github.com/shicai/MobileNet-Caffe} \\
ShuffleNet & \url{https://github.com/msnqqer/ShuffleNet} \\
\bottomrule
\end{tabular}
}
\end{table}

Provided a pre-trained high-precision model, weight scaling factors~$\delta_1^L$ are initialized to cover the dynamic range of the pre-trained weights, i.e., the $99$-th percentile magnitude of the weights in each layer. Similarly, activation scaling factors~$\Delta_1^L$ are set to cover the dynamic range of the activations in each layer, which are obtained by feeding a small number of training data to the pre-trained model.

For quantization of ImageNet classification networks, we employ the Adam optimizer~\citep{kingma2014adam}. The learning rate is set to be $10^{-5}$ and we train $300$k batches with the batch size of $256$, $128$, $32$ and $64$ for AlexNet, ResNet-18, MobileNet and ShuffleNet, respectively. Then, we decrease the learning rate to $10^{-6}$ and train $200$k more batches. For the learnable regularization coefficient~$\lambda$, we let $\lambda=e^{\omega}$ and learn $\omega$ instead in order to make $\lambda$ always positive in training. The initial value of $\omega$ is set to be $0$, and it is updated with the Adam optimizer using the learning rate of $10^{-4}$. For pruning of MobileNet and ShuffleNet, the Adam optimizer is used for $500$k batches with learning rate $10^{-5}$, without decreasing the learning rate to $10^{-6}$ at $300$k batches. The initial value of $\omega$ is set to be $10$ in pruning. The other settings are the same as described above for quantization. Then, pruned MobileNet and ShuffleNet models are quantized by following the same training procedure as described above for quantization. For quantization of image super resolution networks, we train the quantized models using the Adam optimizer for $3$M batches with the batch size of $128$. We use the learning rate of $10^{-5}$. The initial value for $\omega$ is set to be $0$ and it is updated by the Adam optimizer using the learning rate of $10^{-5}$.

\subsection{Experimental results} \label{sec:exp:res}

\subsubsection{AlexNet quantization} \label{sec:exp:res:alexnet}

\setlength{\tabcolsep}{0.4em}
\begin{table}[t!]
\centering
\caption{AlexNet quantization results on ImageNet classification in comparison to DoReFa-Net~\citep{zhou2016dorefa}.\label{sec:exp:res:alexnet:tbl:01}}
{\scriptsize
\begin{tabular}{ccccc}
\toprule
\multirow{2}{*}{\shortstack[c]{Quantized\\layers}} & Weights & Activations & \multicolumn{2}{c}{Top-1 / Top-5 accuracy (\%)} \\
\cmidrule{4-5}
  &                            &            & Ours        & DoReFa-Net~\citep{zhou2016dorefa}* \\
\midrule
Pre-trained model
  & 32-bit FLP                 & 32-bit FLP & \multicolumn{2}{c}{58.0 / 80.8} \\
\cmidrule{1-5}
\multirow{7}{*}{\shortstack[c]{(1) All layers}}
  & 8-bit FXP                  & 8-bit FXP  & 57.7 / 80.5 & 57.6 / 80.8 \\
  & 4-bit FXP                  & 4-bit FXP  & 56.5 / 79.4 & 56.9 / 80.3 \\
  & 2-bit FXP                  & 2-bit FXP  & 53.5 / 77.3 & 43.0 / 68.1 \\
\cmidrule{2-5}
  & \multirow{4}{*}{1-bit FXP} & 8-bit FXP  & 52.2 / 75.8 & 47.5 / 72.1 \\
  &                            & 4-bit FXP  & 52.0 / 75.7 & 45.1 / 69.7 \\
  &                            & 2-bit FXP  & 50.5 / 74.6 & 43.6 / 68.3 \\
  &                            & 1-bit FXP  & 41.1 / 66.6 & 19.3 / 38.2 \\
\cmidrule{1-5}
\multirow{7}{*}{\shortstack[c]{(2) Except the first\\and the last layers}} 
  & 8-bit FXP                  & 8-bit FXP  & 57.7 / 80.6 & 57.5 / 80.7 \\
  & 4-bit FXP                  & 4-bit FXP  & 56.6 / 79.8 & 56.9 / 80.1 \\
  & 2-bit FXP                  & 2-bit FXP  & 54.1 / 77.9 & 53.1 / 77.3 \\
\cmidrule{2-5}
  & \multirow{4}{*}{1-bit FXP} & 8-bit FXP  & 54.8 / 78.1 & 51.2 / 75.5 \\
  &                            & 4-bit FXP  & 54.8 / 78.2 & 51.9 / 75.9 \\
  &                            & 2-bit FXP  & 53.0 / 76.8 & 49.3 / 74.1 \\
  &                            & 1-bit FXP  & 43.9 / 69.0 & 40.2 / 65.5 \\
\bottomrule
\multicolumn{5}{r}{* from our experiments using their code.}
\end{tabular}
}
\end{table}

In Table~\ref{sec:exp:res:alexnet:tbl:01}, we compare our quantization method to DoReFa-Net~\citep{zhou2016dorefa} for the AlexNet model in \citet{krizhevsky2012imagenet}. Since DoReFa-Net does not consider weight pruning, we neither apply pruning here. The DoReFa-Net results in Table~\ref{sec:exp:res:alexnet:tbl:01} are (re-)produced by us from their code\footnote{\url{https://github.com/ppwwyyxx/tensorpack/tree/master/examples/DoReFa-Net}}, and we use the same training hyperparameters and epochs as we described in Section~\ref{sec:exp:setting} for fair comparison. We evaluate two cases where (1) all layers are quantized, and (2) all layers except the first and the last layers are quantized. The results in Table~\ref{sec:exp:res:alexnet:tbl:01} show that 4-bit quantization is needed for accuracy loss less than $1$\%. For binary weights, we observe some accuracy loss of more or less than $10$\%. However, we can see that our quantization scheme performs better than DoReFa-Net in particular for low-precision cases, where the quantization error is larger and the mismatch problem of the forward and backward passes is more severe.

\subsubsection{ResNet-18 quantization} \label{sec:exp:res:resnet18}

\begin{figure*}[t!]
\centering
\includegraphics[width=.88\textwidth]{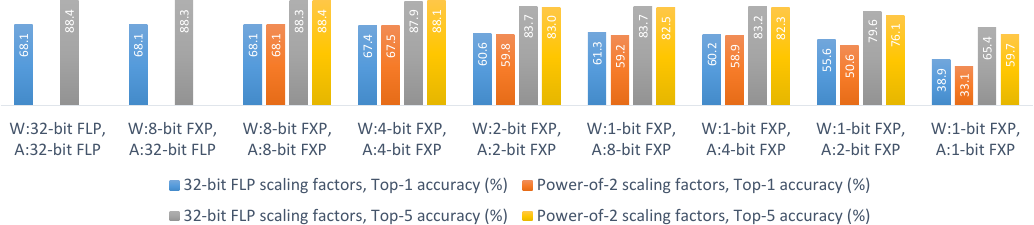}
\caption{Ablation study of ResNet-18 quantization on ImageNet classification. We use ``W: Weight precision'' and ``A: Activation precision'' to denote weight and activation precisions, respectively. FLP and FXP stands for floating-point and fixed-point formats, respectively.\label{sec:exp:res:resnet18:fig:01}}
\end{figure*}

Figure~\ref{sec:exp:res:resnet18:fig:01} presents the accuracy of the low-precision ResNet-18~\citep{he2016deep} models obtained from our quantization method. The experiments on ResNet-18 are mainly for ablation study. In particular, we compare weight and activation quantization for various low-precision settings. The loss due to weight quantization is relatively less than the loss due to activation quantization, which is consistent with the results from DoReFa-Net~\citep{zhou2016dorefa}. We also compare the low-precision models obtained with and without the constraint of power-of-two scaling factors. In fixed-point computations (see Figure~\ref{sec:model:fig:01}), it is more appealing for scaling factors (i.e., quantization bin sizes) to be powers of two so they can be implemented by simple bit-shift, rather than with scalar multiplication. For power-of-two scaling factors, we perform rounding of scaling factors into their closest power-of-two values in the forward pass, while the rounding function is replaced with the identity function in the backward pass. We observe small performance degradation due to the constraint of power-of-two scaling factors in our experiments. 

\setlength{\tabcolsep}{0.4em}
\begin{table}[t!]
\centering
\caption{Accuracy loss comparison of the 4-bit FXP ResNet-18 models. Since the baseline 32-bit FLP model shows different accuracy in each method, we compare the accuracy loss of 4-bit FXP models from 32-bit FLP models.\label{sec:exp:res:resnet18:tbl:01}}
{\scriptsize
\begin{tabular}{ccccccc}
\toprule
Weights    & Activations & \multicolumn{5}{c}{Top-1 accuracy (\%)} \\
\cmidrule(l{2pt}r{2pt}){3-7}
           &             & \multicolumn{2}{c}{Ours} & BCGD & PACT* & DSQ \\
           &             & 1-crop & 10-crop & \cite{yin2019blended} & \cite{choi2018pact} & \cite{gong2019differentiable} \\
\midrule
32-bit FLP & 32-bit FLP  & 68.1 & 69.8         & 69.6 & 70.2 & 69.9 \\
4-bit FXP  & 4-bit FXP   & 67.4 & 69.5         & 67.4 & 69.2 & 69.6 \\
\midrule
\multicolumn{2}{c}{Accuracy (\%) difference}
                         & 0.7  & \textbf{0.3} & 2.2  & 1.0  & \textbf{0.3} \\
\bottomrule
\multicolumn{7}{r}{* The first and the last layers are not quantized.}
\end{tabular}
}
\end{table}

In Table~\ref{sec:exp:res:resnet18:tbl:01}, we compare the proposed quantization scheme to the existing quantization methods from \cite{yin2019blended,choi2018pact,gong2019differentiable} for 4-bit weight and 4-bit activation quantization of ResNet-18. All convolutional and fully-connected layers of ResNet-18 are quantized in \cite{yin2019blended,gong2019differentiable}, and ours, while the first and the last layers are not quantized in \cite{choi2018pact}. Since the baseline 32-bit model shows different accuracy in each method, we compare the accuracy difference between 32-bit floating-point models and 4-bit fixed-point models. For our method, we also show the accuracy obtained by using the average score from 10 different crops of the input (called 10-crop testing), where the baseline accuracy of our 32-bit floating-point model is aligned with the others. The results show that the proposed quantization scheme achieves 4-bit ResNet-18 quantization whose accuracy loss is comparable to the state-of-the-art methods. In particular, the accuracy loss from 4-bit quantization is shown to be very small and less than 1\% in our scheme.

\setlength{\tabcolsep}{0.4em}
\begin{table}[t!]
\centering
\caption{Comparison of learnable and fixed regularization coefficients for ResNet-18 on ImageNet classification.\label{sec:exp:res:resnet18:tbl:02}}
{\scriptsize
\begin{tabular}{cccccc}
\toprule
Weights & Activations & \multicolumn{4}{c}{Top-1 / Top-5 accuracy (\%)} \\
\cmidrule{3-6}
        &             & \shortstack[c]{Learnable\\$\lambda$} & \shortstack[c]{Fixed\\$\lambda=0.05$} & \shortstack[c]{Fixed\\$\lambda=0.5$} & \shortstack[c]{Fixed\\$\lambda=5$} \\
\midrule
32-bit FLP & 32-bit FLP & \multicolumn{4}{c}{68.1 / 88.4} \\
\midrule
\multirow{4}{*}{1-bit FXP}
           & 8-bit FXP  & 61.3 / 83.7 & 60.0 / 83.1 & 60.0 / 83.0 & 57.9 / 81.6 \\
           & 4-bit FXP  & 60.2 / 83.2 & 58.1 / 81.5 & 57.4 / 81.1 & 58.6 / 82.2 \\
           & 2-bit FXP  & 55.6 / 79.6 & 53.5 / 78.2 & 52.9 / 77.8 & 53.1 / 78.1 \\
           & 1-bit FXP  & 38.9 / 65.4 & 37.0 / 63.4 & 36.5 / 63.1 & 37.0 / 63.1 \\
\bottomrule
\end{tabular}
}
\end{table}

\textbf{Learnable versus fixed regularization coefficients}. In Table~\ref{sec:exp:res:resnet18:tbl:02}, we compare the performance of quantized ResNet-18~\citep{he2016deep} models when we use learnable and fixed regularization coefficients, respectively. Observe that the proposed learnable regularization method outperforms the conventional regularization method with a fixed coefficient in various low-precision settings.

\begin{figure}[t!]
\centering
\subfigure[Learnable $\lambda$]{\includegraphics[width=0.49\columnwidth]{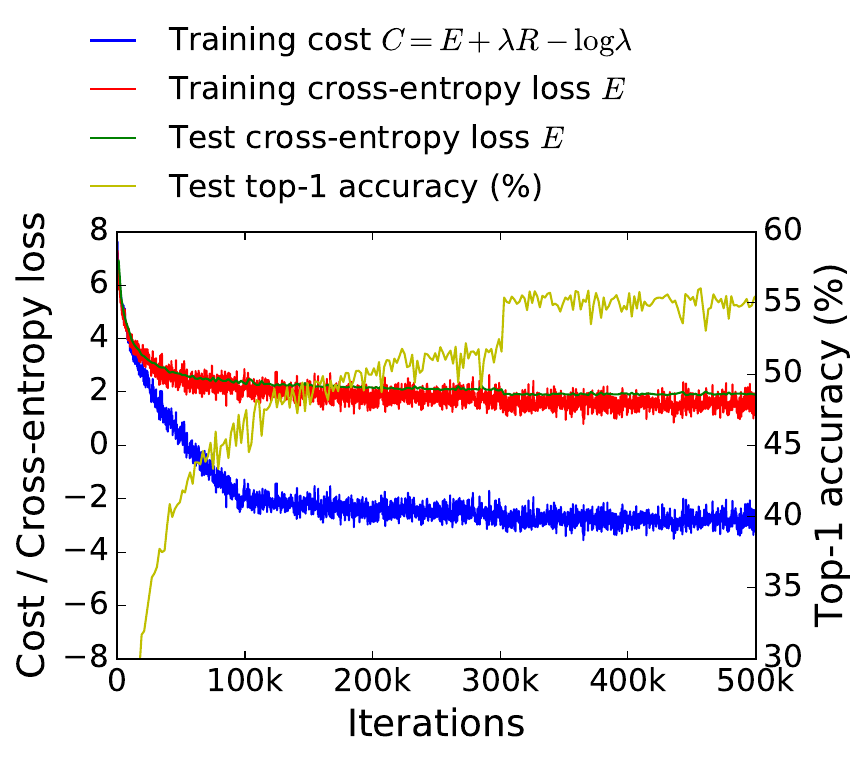}\quad\includegraphics[width=0.49\columnwidth]{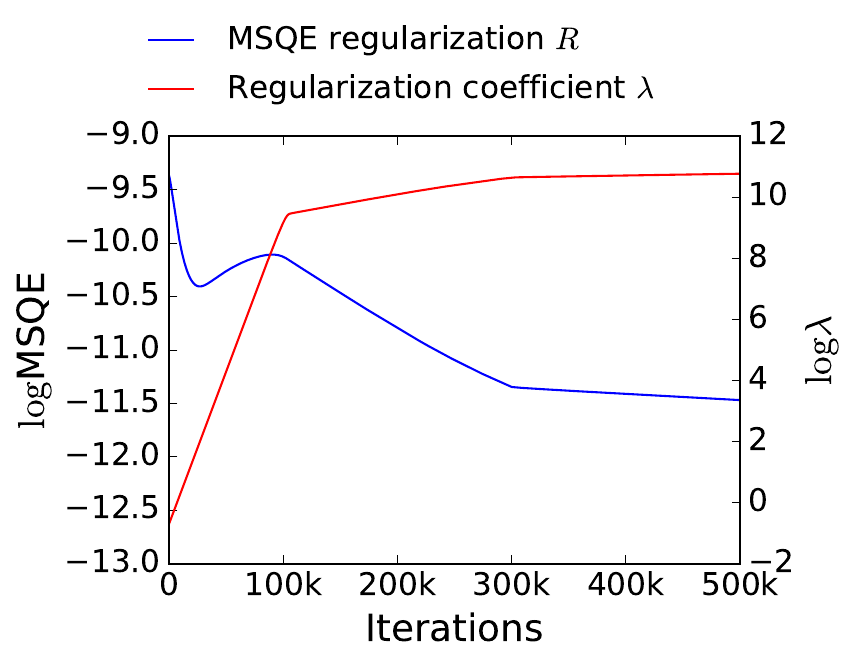}}\\
\subfigure[Fixed $\lambda=0.5$]{\includegraphics[width=0.49\columnwidth]{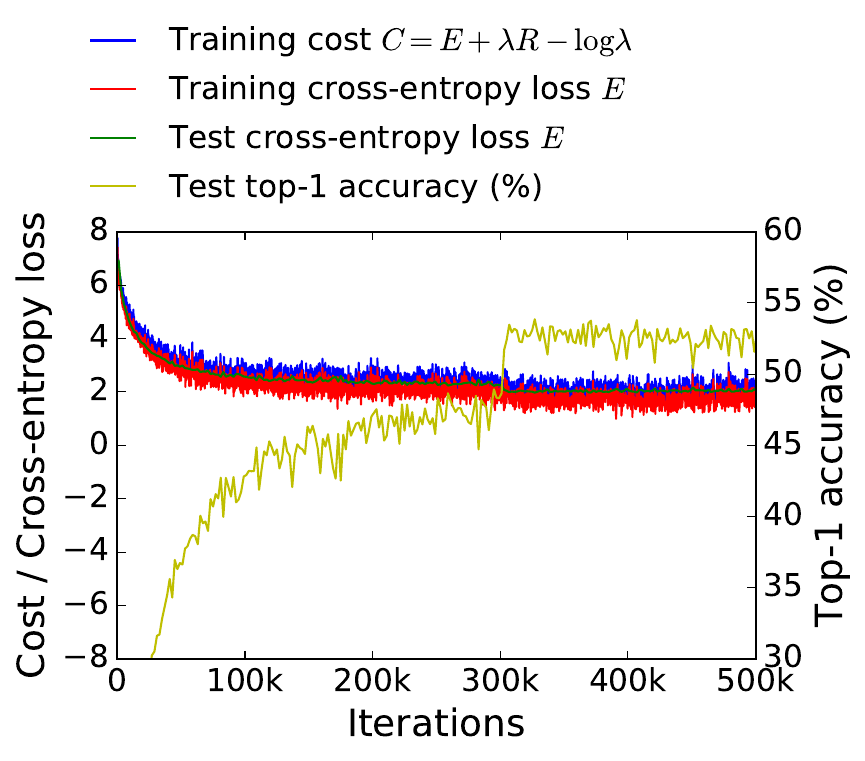}\quad\includegraphics[width=0.49\columnwidth]{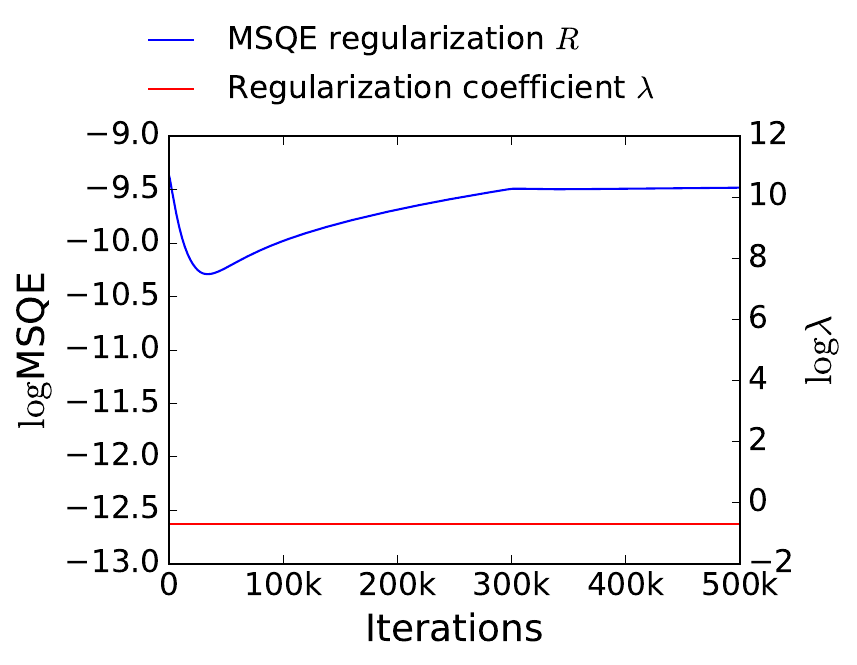}}
\caption{ResNet-18 model training convergence curves for binary weights and 2-bit activations. We compare the convergence curves with learnable and fixed regularization coefficients.\label{sec:exp:res:resnet18:fig:02}}
\end{figure}

In Figure~\ref{sec:exp:res:resnet18:fig:02}, we compare the convergence curves when learnable and fixed regularization coefficients are used, respectively. Using a learnable regularization coefficient, the MSQE regularization term decreases (although there is a bump in the middle) while $\lambda$ increases in training. However, using a fixed regularization coefficient, the MSQE regularization term saturates and even increases after some point as training goes on, which implies that the mismatch of the forward and backward passes is not resolved. The unresolved mismatch eventually turns into accuracy loss, as shown in the figure.

\subsubsection{MobileNet and ShuffleNet compression} \label{sec:exp:res:comp}

\setlength{\tabcolsep}{.4em}
\begin{table*}[t!]
\centering
\caption{Low-precision MobileNet and ShuffleNet compression results for ImageNet classification. For ablation study, we compare pruning-only results and pruning+quantization results with various low-precision setting. We also show the compression results with and without entropy coding, where we used bzip2 as a specific entropy coding scheme.\label{sec:exp:res:comp:tbl:01}}
{\scriptsize
\begin{tabular}{ccccccc}
\toprule
Method & Weights & Activations & \multicolumn{2}{c}{MobileNet v1} & \multicolumn{2}{c}{ShuffleNet} \\
\cmidrule(l{2pt}r{2pt}){4-5} \cmidrule(l{2pt}r{2pt}){6-7}
 & & & Top-1 / Top-5 & Compression ratio     & Top-1 / Top-5 & Compression ratio \\
 & & & accuracy (\%) & with / without bzip2  & accuracy (\%) & with / without bzip2 \\
\midrule
Pre-trained model
  & 32-bit FLP & 32-bit FLP                   & 70.9 / 89.9 & -    & 65.4 / 86.4 & -\\
\cmidrule{1-7}
Ours: pruning (50\%)
  & \multirow{3}{*}{32-bit FLP}
               & \multirow{3}{*}{32-bit FLP}  & 70.2 / 89.7 & ~2.01 / 1.00 & 65.3 / 86.4 & 1.99 / 1.00 \\
\newlength{\myl}\settowidth{\myl}{Ours: }\hspace{\the\myl}pruning (55\%)
  &            &                              & 70.0 / 89.5 & ~2.22 / 1.00 & 64.7 / 86.0 & 2.20 / 1.00 \\
\settowidth{\myl}{Ours: }\hspace{\the\myl}pruning (60\%)
  &            &                              & 69.5 / 89.3 & ~2.49 / 1.00 & 63.6 / 85.5 & 2.45 / 1.00 \\
\cmidrule{1-7}
\multirow{11}{*}{\shortstack[c]{Ours: pruning (50\%)\\+ quantization}}
  & 8-bit FXP  & 8-bit FXP                    & 70.8 / 90.1 & ~4.83 / 4.00 & 65.8 / 86.7 & 4.99 / 4.00 \\
  & 6-bit FXP  & 6-bit FXP                    & 70.5 / 89.9 & ~6.11 / 5.33 & 65.7 / 86.7 & 5.81 / 5.33 \\
  & 5-bit FXP  & 5-bit FXP                    & 69.7 / 89.3 & ~7.13 / 6.40 & 64.0 / 85.6 & 6.78 / 6.40 \\
  & 4-bit FXP  & 4-bit FXP                    & 66.9 / 87.7 & ~9.87 / 8.00 & 59.5 / 82.6 & 9.59 / 8.00 \\
\cmidrule{2-7}
  & 6-bit FXP  & \multirow{3}{*}{8-bit FXP}   & 70.6 / 90.0 & ~6.11 / 5.33 & 66.3 / 87.1 & 5.81 / 5.33 \\
  & 5-bit FXP  &                              & 70.3 / 89.7 & ~7.13 / 6.40 & 65.8 / 86.7 & 6.79 / 6.40 \\
  & 4-bit FXP  &                              & 69.7 / 89.2 & ~8.65 / 8.00 & 64.8 / 86.2 & 8.26 / 8.00 \\
\cmidrule{2-7}
  & 6-bit FXP  & \multirow{3}{*}{32-bit FLP}  & 70.7 / 90.0 & ~6.12 / 5.33 & 66.3 / 87.1 & 5.81 / 5.33 \\
  & 5-bit FXP  &                              & 70.4 / 89.8 & ~7.13 / 6.40 & 65.8 / 86.9 & 6.78 / 6.40 \\
  & 4-bit FXP  &                              & 69.3 / 89.0 & 10.01 / 8.00 & 64.1 / 85.8 & 9.71 / 8.00 \\
\cmidrule{1-7}
Tensorflow 8-bit model*
  & 8-bit FXP  & 8-bit FXP                    & 70.1 / 88.9 & ~~N/A / 4.00 & N/A         & N/A  \\
\cmidrule{1-7}
\multirow{3}{*}{\shortstack[c]{Relaxed\\quantization~\cite{louizos2018relaxed}}}
  & 8-bit FXP  & 8-bit FXP                    & 70.4 / 89.4 & ~~N/A / 4.00 & N/A         & N/A \\
  & 6-bit FXP  & 6-bit FXP                    & 68.0 / 88.0 & ~~N/A / 5.33 & N/A         & N/A \\
  & 5-bit FXP  & 5-bit FXP                    & 61.4 / 83.7 & ~~N/A / 6.40 & N/A         & N/A \\
\bottomrule
\multicolumn{7}{r}{* \url{https://github.com/tensorflow/models/blob/master/research/slim/nets/mobilenet_v1.md}}
\end{tabular}
}
\end{table*}
\begin{figure*}[t!]
\centering
\includegraphics[width=.88\textwidth]{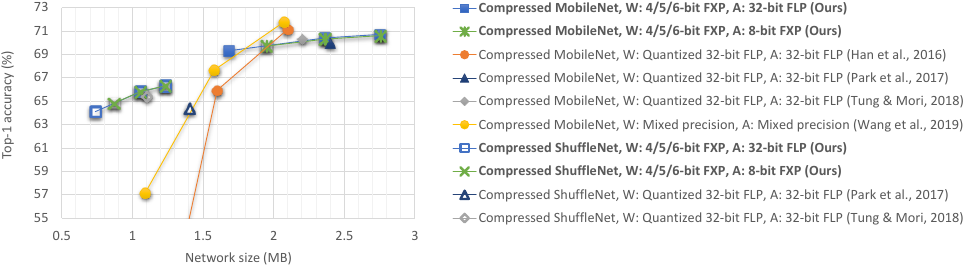}
\caption{Comparison of our low-precision MobileNet and ShuffleNet compression results to the ones of the state-of-the-art network compression methods on ImageNet classification. We use ``W: Weight precision'' and ``A: Activation precision'' to denote weight and activation precisions used in the compressed models, respectively.\label{sec:exp:res:comp:fig:01}}
\end{figure*}
\setlength{\tabcolsep}{0.4em}

We mainly evaluate our method to obtain compressed low-precision MobileNet~\citep{howard2017mobilenets} and ShuffleNet~\citep{zhang2018shufflenet} models for ImageNet classification. MobileNet and ShuffleNet are state-of-the-art ImageNet classification networks developed for efficient inference on resource-limited platforms. Compression and quantization of such efficient networks are important in practice to lower latency and to improve power-efficiency further in mobile and edge devices. It is typically more difficult to compress and quantize such networks of efficient architectures. For MobileNet and ShuffleNet compression, we prune $50$\% weights from their pre-trained models as described in Section~\ref{sec:reg:pruning} so that the accuracy loss due to pruning is marginal. Then, we employ our weight and activation quantization method. After converting into sparse low-precision models, universal source coding with bzip2~\citep{seward1998bzip2} follows to compress the fixed-point low-precision weights.

In Table~\ref{sec:exp:res:comp:tbl:01}, for ablation study, we compare pruning-only results and pruning+quantization results with various low-precision setting. We also show the compression results with and without entropy coding, where we use bzip2 as a specific entropy coding scheme. Observe that the accuracy loss is marginal when we prune 50\% weights for both MobileNet and ShuffleNet. After pruning 50\% weights, we quantize the pruned models. Similar to the AlexNet and ResNet-18 results, the accuracy loss from quantization is more severe when we decrease the activation bit-width than the weight bit-width. From the experiments, we obtain low-precision models of 5-bit weights and 8-bit activations with top-1 accuracy loss of 0.6\% only. The compression ratio of these low-precision models is 6.40 without bzip2 compression, but it increases and becomes 7.13 and 6.79 for MobileNet and ShuffleNet, respectively, after bzip2 compression. We also show that our scheme outperforms the existing quantization schemes from tensorflow and \cite{louizos2018relaxed}.

In Figure~\ref{sec:exp:res:comp:fig:01}, we compare the compression ratios of our scheme and the existing network compression methods in \citet{han2015deep,park2017weighted,tung2018deep,wang2019haq}. Our low-precision network compression scheme shows comparable compression ratios to the state-of-the-art weight compression schemes. We emphasize that our scheme produces low-precision models of fixed-point weights and activations that support efficient inference of fixed-point operations, while the previous compression schemes, except \citet{wang2019haq}, produces quantized weights that are still floating-point numbers and thus floating-point operations are necessary to achieve the presented accuracy of them. The hardware-aware automated quantization in \citet{wang2019haq} achieved impressive compression results by searching for a quantized model of ``mixed'' precision for different layers with reinforcement learning, but not all hardware supports mixed precision operations.

\subsubsection{Image super resolution network quantization} \label{sec:exp:res:sr}

\begin{table*}[t!]
\centering
\caption{CT-SRCNN (9-layer) quantization results for upscaling factor~$3$.\label{sec:exp:res:sr:tbl:01}}
{\scriptsize
\begin{tabular}{cccccccc}
\toprule
Model & Method & Weights & Activations & Set-14 PSNR (dB) & Set-14 SSIM & PSNR (dB) loss & SSIM loss \\
\midrule
\multirow{7}{*}{\shortstack[c]{SRCNN\\3-layer}}
	& Pre-trained model						& 32-bit FLP & 32-bit FLP                 & 29.05 & 0.8161 & -    & -	   \\
\cmidrule{2-8}
	& \multirow{4}{*}{Ours}					& 8-bit FXP  & \multirow{4}{*}{8-bit FXP} & 29.03 & 0.8141 & 0.02 & 0.0020 \\
	& 										& 4-bit FXP  &                            & 28.99 & 0.8133 & 0.06 & 0.0028 \\
	& 										& 2-bit FXP  &                            & 28.72 & 0.8075 & 0.33 & 0.0086 \\
	& 										& 1-bit FXP  &                            & 28.53 & 0.8000 & 0.52 & 0.0161 \\
\cmidrule{2-8}
	& Ristretto~\citet{gysel2018ristretto}*
											& 8-bit FXP  & 8-bit FXP                  & 28.58 & 0.7827 & 0.46 & 0.0328 \\
\midrule
\multirow{7}{*}{\shortstack[c]{CT-SRCNN\\5-layer}}
	& Pre-trained model						& 32-bit FLP & 32-bit FLP                 & 29.56 & 0.8273 & -    & -	   \\
\cmidrule{2-8}
	& \multirow{4}{*}{Ours}					& 8-bit FXP  & \multirow{4}{*}{8-bit FXP} & 29.54 & 0.8267 & 0.02 & 0.0006 \\
	& 										& 4-bit FXP  &                            & 29.48 & 0.8258 & 0.08 & 0.0015 \\
	& 										& 2-bit FXP  &                            & 29.28 & 0.8201 & 0.28 & 0.0072 \\
	& 										& 1-bit FXP  &                            & 29.09 & 0.8171 & 0.47 & 0.0102 \\
\cmidrule{2-8}
	& Ristretto~\citet{gysel2018ristretto}*
											& 8-bit FXP  & 8-bit FXP                  & 29.04 & 0.8111 & 0.53 & 0.0148 \\
\midrule
\multirow{7}{*}{\shortstack[c]{CT-SRCNN\\9-layer}}
	& Pre-trained model						& 32-bit FLP & 32-bit FLP                 & 29.71 & 0.8300 & -    & -	   \\
\cmidrule{2-8}
	& \multirow{4}{*}{Ours}					& 8-bit FXP  & \multirow{4}{*}{8-bit FXP} & 29.67 & 0.8288 & 0.04 & 0.0012 \\
	& 										& 4-bit FXP  &                            & 29.63 & 0.8285 & 0.08 & 0.0015 \\
	& 										& 2-bit FXP  &                            & 29.37 & 0.8236 & 0.34 & 0.0064 \\
	& 										& 1-bit FXP  &                            & 29.20 & 0.8193 & 0.51 & 0.0107 \\
\cmidrule{2-8}
	& Ristretto~\citet{gysel2018ristretto}*
											& 8-bit FXP  & 8-bit FXP                  & 29.05 & 0.8065 & 0.74 & 0.0234 \\
\midrule
Bicubic
	& -										& -          & -                          & 27.54 & 0.7742 & -    & -      \\
\bottomrule
\multicolumn{8}{r}{* from our experiments using their code at \url{https://github.com/pmgysel/caffe}.}
\end{tabular}
}
\end{table*}

The image super resolution problem is to synthesize a high-resolution image from a low-resolution one. The DNN output is the high-resolution image corresponding to the input low-resolution image, and thus the loss due to quantization is more prominent. We evaluate the proposed method on SRCNN~\citep{dong2016image} and cascade-trained SRCNN (CT-SRCNN)~\citep{ren2018ctsrcnn} for image super resolution. The objective image quality metric measured by the peak signal-to-noise ratio (PSNR) and the perceptual score measured by the structural similarity index (SSIM)~\citep{wang2004image} are compared for Set-14 image dataset~\citep{zeyde2010single} in Table~\ref{sec:exp:res:sr:tbl:01} for 3-layer SRCNN, 5-layer CT-SRCNN, and 9-layer CT-SRCNN, respectively. Observe that our method successfully yields low-precision models of 8-bit weights and activations at negligible loss, and they are better than the results that we obtain with one of the previous works, Ristretto~\citep{gysel2018ristretto}. It is interesting to see that the PSNR loss of using binary weights and 8-bit activations is $0.5$~dB only.

\section{Conclusion} \label{sec:conclusion}

In this paper, we proposed a method to quantize deep neural networks (DNNs) by regularization to produce low-precision DNNs for efficient fixed-point inference. Although our training happens in high precision particularly for its backward passes and gradient descent, its forward passes use quantized low-precision weights and activations, and thus the resulting networks can be operated on low-precision fixed-point hardware at inference time. The proposed scheme alleviates the mismatch problem in the forward and backward passes of low-precision network training by using MSQE regularization. Moreover, we proposed a novel learnable regularization coefficient to reinforce the convergence of high-precision weights to their quantized values when using MSQE regularization. We also discussed how a similar regularization technique can be employed for weight pruning with partial L2 regularization.

We showed by experiments that the proposed quantization algorithm successfully produces low-precision DNNs of binary weights for classification problems, such as ImageNet classification, as well as for regression and image synthesis problems, such as image super resolution. For MobileNet and ShuffleNet compression, we obtained sparse (50\% weights are pruned) low-precision models of 5-bit weights and 8-bit activations with compression ratios of $7.13$ and $6.79$, respectively, at marginal accuracy loss. For image super resolution, we only lost $0.04$ dB PSNR when using 8-bit weights and activations, instead of 32-bit floating-point numbers. 

\appendix

\section{Gradient Descent} \label{app:01}

\subsection{Gradients for weights} \label{sec:train:weight}

The gradient of the cost function~$C_n$ in \eqref{sec:reg:weightquant:eq:03} for $w$ satisfies
\begin{equation} \label{sec:train:weight:eq:01}
\nabla_wC_n
=\nabla_wE+\lambda\nabla_wR_n,
\end{equation}
for weight~$w$ of layer~$l$, $1\leq l\leq L$. The first partial derivative in the right side of \eqref{sec:train:weight:eq:01} can be obtained efficiently by the backpropagation algorithm. For backpropagation through the weight quantization function, we adopt the following approximation similar to straight-through estimator~\citep{bengio2013estimating}:
\begin{equation} \label{sec:train:weight:eq:03}
\nabla_wQ_n(w;\delta_l) \\
\triangleq\begin{cases}
1_{\frac{w}{\delta_l}\in[-2^{n-1}-\frac{1}{2},2^{n-1}-\frac{1}{2}]}, & n>1, \\
1_{\frac{w}{\delta_l}\in[-2,2]}, & n=1,
\end{cases}
\end{equation}
where $1_\mathcal{E}$ is an indication function such that it is one if $\mathcal{E}$ is true and zero otherwise. Namely, we pass the gradient through the quantization function when the weight is within the clipping boundary. To give some room for the weight to move around the boundary in stochastic gradient descent, we additionally allow some margin of $\delta_l/2$ for $n\geq2$ and $\delta_l$ for $n=1$. Outside the clipping boundary with some margin, we pass zero.

For weight~$w$ of layer~$l$, $1\leq l\leq L$, the partial derivative of the regularizer~$R_n$ satisfies
\begin{equation} \label{sec:train:weight:eq:04}
\nabla_wR_n
=\frac{2}{N}(w-Q_n(w;\delta_l)),
\end{equation}
almost everywhere except some non-differentiable points of $w$ at quantization bin boundaries~$\mathcal{U}_n(\delta_l)$ given by
\begin{equation} \label{sec:train:weight:eq:05}
\mathcal{U}_n(\delta_l)
=\left\{\frac{2i+1-2^n}{2}\delta_l,i=0,1,\dots,2^n-2\right\},
\end{equation}
for $n>1$ and $\mathcal{U}_1(\delta_l)=\left\{0\right\}$. If the weight is located at one of these boundaries, it actually makes no difference to update $w$ to either direction of $w-\epsilon$ or $w+\epsilon$, in terms of its quantization error. Thus, we let
\begin{equation} \label{sec:train:weight:eq:06}
\nabla_wR_n\triangleq0,
\ \ \
\text{if $w\in\mathcal{U}_n(\delta_l)$}.
\end{equation}
From \eqref{sec:train:weight:eq:01}--\eqref{sec:train:weight:eq:06}, we finally have
\[
\nabla_wC_n
=\nabla_wE
+\frac{2\lambda}{N}(w-Q_n(w;\delta_l))1_{w\notin\mathcal{U}_n(\delta_l)}.
\]
\begin{remark} \label{sec:train:weight:remark:01}
If the weight is located at one of the bin boundaries, the weight gradient is solely determined by the network loss function derivative and thus the weight is updated towards the direction to minimize the network loss function. Otherwise, the regularization term impacts the gradient as well and encourages the weight to converge to the closest bin center as far as the loss function changes small. The regularization coefficient trades off these two contributions of the network loss function and the regularization term.
\end{remark}

\subsection{Gradient for the regularization coefficient} \label{sec:train:coeff}

The gradient of the cost function for $\lambda$ is given by
\begin{equation} \label{sec:train:coeff:eq:01}
\nabla_\lambda C_n
=R_n(\mathcal{W}_1^L;\delta_1^L)-\frac{1}{\lambda}.
\end{equation}
Observe that $\lambda$ tends to $1/R_n$ in gradient descent.
\begin{remark} \label{sec:train:coeff:remark:01}
Recall that weights gradually tend to their closest quantization output levels to reduce the regularizer~$R_n$ (see Remark~\ref{sec:train:weight:remark:01}). As the regularizer~$R_n$ decreases, the regularization coefficient~$\lambda$ gets larger by gradient descent using \eqref{sec:train:coeff:eq:01}. Then, a larger regularization coefficient further forces weights to move towards quantized values in the following update. In this manner, weights gradually converges to quantized values.
\end{remark}

\subsection{Gradients for scaling factors} \label{sec:train:scale}

For scaling factor optimization, we approximately consider the MSQE regularization term only for simplicity. Using the chain rule, it follows that
\begin{equation} \label{sec:train:scale:eq:01}
\begin{split}
\nabla_{\delta_l}C_n
&\approx\nabla_{\delta_l}R_n \\
&=-\frac{2\lambda}{N}\sum_{w\in\mathcal{W}_l}(w-Q_n(w;\delta_l))\nabla_{\delta_l}Q_n(w;\delta_l),
\end{split}
\end{equation}
for $1\leq l\leq L$. Moreover, it can be shown that
\begin{equation} \label{sec:train:scale:eq:02}
\begin{split}
\nabla_{\delta_l}Q_n(w;\delta_l)
&=r_n(w;\delta_l) \\
&\triangleq\begin{cases}
\clip_n(\round(w/\delta_l)), & n>1, \\
\sign(w), & n=1,
\end{cases}
\end{split}
\end{equation}
almost everywhere except some non-differentiable points of $\delta_l$ satisfying
\begin{equation} \label{sec:train:scale:eq:03}
\frac{w}{\delta_l}\in\left\{\frac{2i+1-2^n}{2},i=0,1,\dots,2^n-2\right\},
\end{equation}
for $n>1$.
Similar to \eqref{sec:train:weight:eq:06}, we let
\begin{equation} \label{sec:train:scale:eq:04}
\nabla_{\delta_l}Q_n(w;\delta_l)\triangleq0,
\ \ \
\text{if $w\in\mathcal{U}_n(\delta_l)$},
\end{equation}
so that the scaling factor~$\delta_l$ is not impacted by the weights at the bin boundaries. From \eqref{sec:train:scale:eq:01}--\eqref{sec:train:scale:eq:04}, it follows that
\begin{equation*}
\nabla_{\delta_l}C_n
\approx-\frac{2\lambda}{N}\sum_{w\in\mathcal{W}_l}(w-Q_n(w;\delta_l))r_n(w;\delta_l)1_{w\notin\mathcal{U}_n(\delta_l)}.
\end{equation*}
Similarly, one can derive the gradients for activation scaling factors~$\Delta_0^L$, which we omit here.

\bibliographystyle{IEEEtranS}
\bibliography{ref}

\end{document}